%% file: main.tex
\pdfoutput=1

\documentclass[11pt]{article}

\usepackage{acl}

\usepackage{times}
\usepackage{latexsym}

\usepackage[T1]{fontenc}

\usepackage[utf8]{inputenc}

\usepackage{microtype}

\title{LinkBERT: Pretraining Language Models with Document Links}

\author{Michihiro Yasunaga \quad Jure Leskovec\thanks{~~~Equal senior authorship.} \quad Percy Liang\footnotemark[\value{footnote}]\\
Stanford University\\
\scalebox{0.87}[0.9]{{\tt \{myasu,jure,pliang\}@cs.stanford.edu}}}

\usepackage{amsmath}
\usepackage{amsfonts}
\usepackage{lipsum} 
\usepackage{algorithm}
\usepackage{graphicx}
\usepackage{transparent}
\usepackage{soul}
\usepackage{calc}
\definecolor{darkred}{HTML}{bb0000}
\definecolor{myblue}{HTML}{0096c5} 
\definecolor{RoyalBlue}{HTML}{0272bb}
\definecolor{mypurple}{HTML}{8700e0} 
\definecolor{myorange}{HTML}{d84800} 
\definecolor{ForestGreen}{RGB}{34,139,34}
\usepackage[moderate]{savetrees} 

\usepackage{multirow}
\usepackage{makecell}
\usepackage{arydshln}
\usepackage{colortbl}
\usepackage{booktabs} 

\renewcommand\ttdefault{cmtt}
\usepackage{xspace}
\input{macros}

\usepackage{mathtools}

\renewcommand\ttdefault{cmtt}
\hypersetup{
  colorlinks   = true, 
  urlcolor     = ForestGreen, 
  linkcolor    = darkblue, 
  citecolor   = darkblue 
}

\newcommand{\methodname}{LinkBERT\xspace}

\begin{document}
\setlength{\abovedisplayskip}{6pt}
\setlength{\belowdisplayskip}{6pt}

\maketitle

\input{000_abstract}
\input{010_introduction}
\input{060_related}
\input{020_preliminary}
\input{030_method}
\input{040_experiment}
\input{050_biomed}
\input{070_conclusion}

\section*{Reproducibility}
\renewcommand\ttdefault{cmtt}
Pretrained models, code and data are available at\\
\url{https://github.com/michiyasunaga/LinkBERT}.\\
Experiments are available at\\
\url{https://worksheets.codalab.org/worksheets/0x7a6ab9c8d06a41d191335b270da2902e}.

\section*{Acknowledgment}
We thank Siddharth Karamcheti, members of the Stanford P-Lambda, SNAP and NLP groups, as well as our anonymous reviewers for valuable feedback.
We gratefully acknowledge the support of a PECASE Award; DARPA under Nos. HR00112190039 (TAMI), N660011924033 (MCS); Funai Foundation Fellowship; Microsoft Research PhD Fellowship; 
ARO under Nos. W911NF-16-1-0342 (MURI), W911NF-16-1-0171 (DURIP); NSF under Nos. OAC-1835598 (CINES), OAC-1934578 (HDR), CCF-1918940 (Expeditions), IIS-2030477 (RAPID),
NIH under No. R56LM013365;
Stanford Data Science Initiative, 
Wu Tsai Neurosciences Institute,
Chan Zuckerberg Biohub,
Amazon, JPMorgan Chase, Docomo, Hitachi, Intel, KDDI, Toshiba, NEC, Juniper, and UnitedHealth Group.


\bibliography{main}
\bibliographystyle{acl_natbib}

\input{080_appendix}

\end{document}

%% file: macros.tex
\usepackage{amsmath,amsfonts,bm}

\newcommand{\heading}[1]{\vspace{1mm}\noindent\textbf{#1}~~}

\newcommand\mask{\texttt{[MASK]}\xspace}
\newcommand\cls{\texttt{[CLS]}\xspace}
\newcommand\sep{\texttt{[SEP]}\xspace}
\newcommand{\fenc}{f_\text{enc}}
\newcommand{\fhead}{f_\text{head}}
\newcommand{\basemodel}{$_\texttt{base}$\xspace}
\newcommand{\tinymodel}{$_\texttt{tiny}$\xspace}
\newcommand{\largemodel}{$_\texttt{large}$\xspace}

\def\eqref#1{equation~\ref{#1}}

\def\1{\bm{1}}

\DeclareMathAlphabet{\mathsfit}{\encodingdefault}{\sfdefault}{m}{sl}
\SetMathAlphabet{\mathsfit}{bold}{\encodingdefault}{\sfdefault}{bx}{n}



%% file: 000_abstract.tex
\begin{abstract}
Language model (LM) pretraining can learn various knowledge from text corpora, helping downstream tasks. 
However, existing methods such as BERT model a single document, and do not capture dependencies or knowledge that span across documents.
In this work, we propose \textit{LinkBERT}, an LM pretraining method that leverages links between documents, e.g., hyperlinks. Given a text corpus, we view it as a graph of documents and create LM inputs by placing linked documents in the same context. We then pretrain the LM with two joint self-supervised objectives: masked language modeling and our new proposal, document relation prediction.
We show that LinkBERT outperforms BERT on various downstream tasks across two domains: the general domain (pretrained on Wikipedia with hyperlinks) and biomedical domain (pretrained on PubMed with citation links).
LinkBERT is especially effective for multi-hop reasoning and few-shot QA (+5\% absolute improvement on HotpotQA and TriviaQA), and our biomedical LinkBERT sets new states of the art on various BioNLP tasks (+7\% on BioASQ and USMLE).
We release our pretrained models, \textit{LinkBERT} and \textit{BioLinkBERT}, as well as code and data.\footnote{Available at \url{https://github.com/michiyasunaga/LinkBERT}.}
\vspace{2mm}
\end{abstract}

%% file: 010_introduction.tex
\section{Introduction}\label{sec:intro}

Pretrained language models (LMs), like BERT and GPTs \cite{devlin2018bert, brown2020language}, have shown remarkable performance on many natural language processing (NLP) tasks, such as text classification and question answering, becoming the foundation of modern NLP systems \cite{bommasani2021opportunities}. By performing self-supervised learning, such as masked language modeling \cite{devlin2018bert}, LMs learn to encode various knowledge from text corpora and produce informative representations for downstream tasks \cite{petroni2019language, Bosselut2019COMETCT, t5}.

\begin{figure}[!t]
    \centering
    \vspace{-2mm}
    \hspace{-1mm}
    \includegraphics[width=0.46\textwidth]{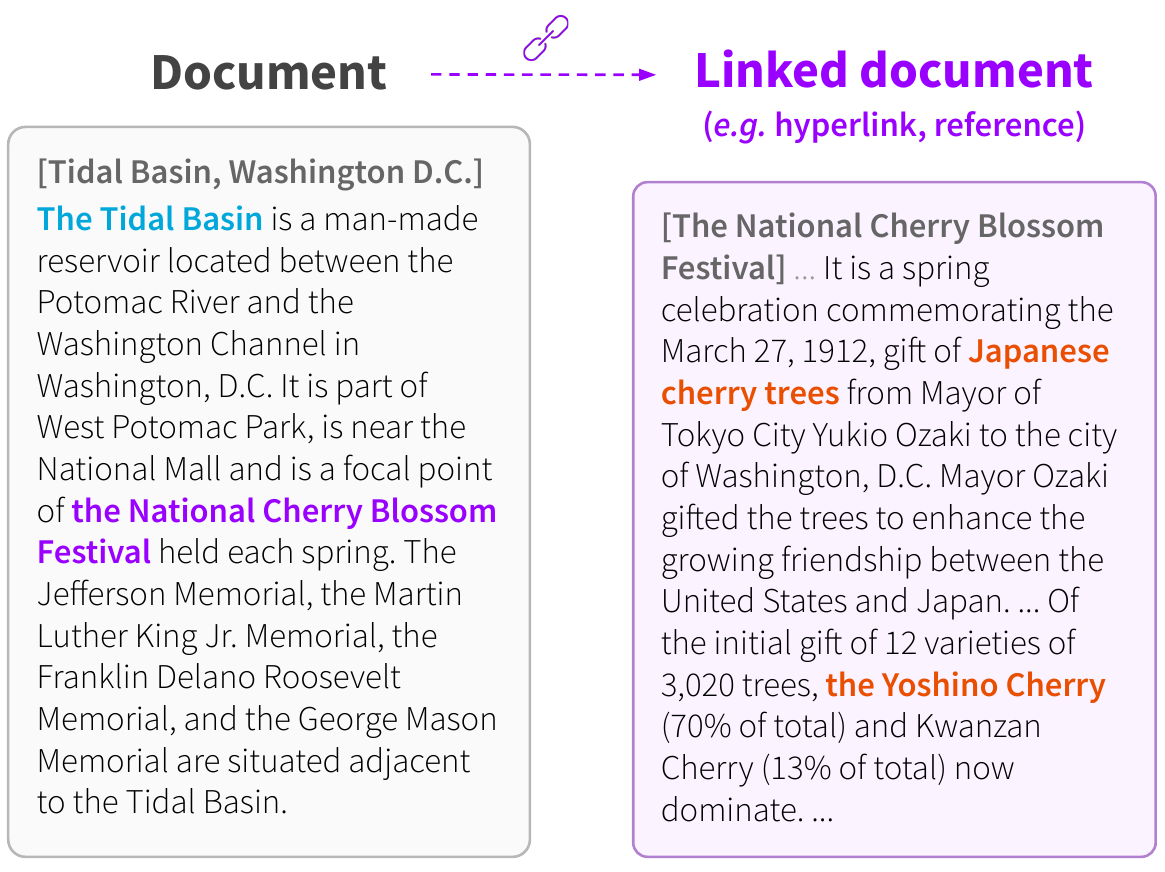}\vspace{-2mm}
    \caption{\small
    Document links (e.g.~hyperlinks) can provide salient multi-hop knowledge. For instance, the Wikipedia article ``\textcolor{myblue}{{Tidal Basin}}'' (left) describes that the basin hosts ``\textcolor{mypurple}{{National Cherry Blossom Festival}}''. The hyperlinked article (right) reveals that the festival celebrates ``\textcolor{myorange}{{Japanese cherry trees}}''. Taken together, the link suggests new knowledge not available in a single document (e.g.~``\textcolor{myblue}{Tidal Basin} has \textcolor{myorange}{Japanese cherry trees}''), which can be useful for various applications, including answering a question ``{What trees can you see at Tidal Basin?}''. We aim to leverage document links to incorporate more knowledge into language model pretraining.
    }
    \vspace{-1mm}
  \label{fig:motivation}
\end{figure}

\begin{figure*}[!th]
    \vspace{-5mm}
    \centering 
    \includegraphics[width=0.99\textwidth]{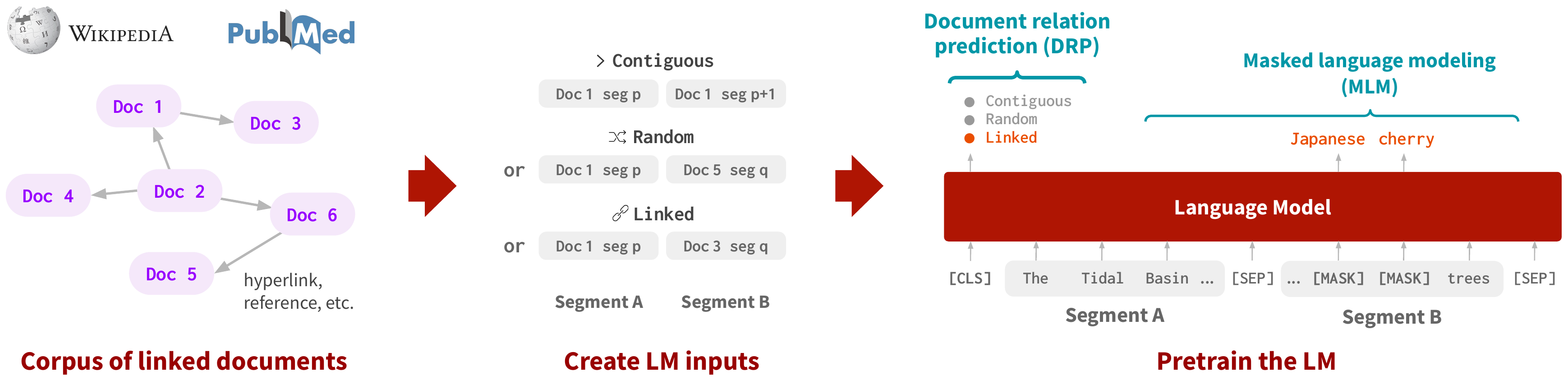}
    \vspace{-1mm}
    \caption{\small
    \textbf{Overview of our approach, LinkBERT}. Given a pretraining corpus, we view it as a graph of documents, with links such as hyperlinks (\S \ref{sec:method-graph}). To incorporate the document link knowledge into LM pretraining, we create LM inputs by placing a pair of linked documents in the same context (\textit{linked}), besides the existing options of placing a single document (\textit{contiguous}) or a pair of random documents (\textit{random}) as in BERT. We then train the LM with two self-supervised objectives: masked language modeling (MLM), which predicts masked tokens in the input, and document relation prediction (DRP), which classifies the relation of the two text segments in the input (\textit{contiguous}, \textit{random}, or \textit{linked}) (\S \ref{sec:method-pretrain}).
    }\label{fig:overview}
    \vspace{-1mm}
\end{figure*}

However, existing LM pretraining methods typically consider text from a single document in each input context \cite{liu2019roberta,joshi2020spanbert} and do not model links between documents. This can pose limitations because documents often have rich dependencies (e.g.~hyperlinks, references), and knowledge can span \textit{across} documents. As an example, in Figure \ref{fig:motivation}, the Wikipedia article ``\textcolor{myblue}{{Tidal Basin, Washington D.C.}}'' (left) describes that the basin hosts ``\textcolor{mypurple}{{National Cherry Blossom Festival}}'', and the hyperlinked article (right) reveals the background that the festival celebrates ``\textcolor{myorange}{{Japanese cherry trees}}''. Taken together, the hyperlink offers new, multi-hop knowledge ``{\textcolor{myblue}{Tidal Basin} has \textcolor{myorange}{Japanese cherry trees}}'', which is not available in the single article ``{Tidal Basin}'' alone. 
Acquiring such multi-hop knowledge in pretraining could be useful for various applications including question answering.
In fact, document links like hyperlinks and references are ubiquitous (e.g.~web, books, scientific literature), and guide how we humans acquire knowledge and even make discoveries \cite{margolis1999concepts}.

In this work, we propose \textit{LinkBERT}, an effective language model pretraining method that incorporates document link knowledge.
Given a text corpus, we obtain links between documents such as hyperlinks, and create LM inputs by placing linked documents in the same context, besides the existing option of placing a single document or random documents as in BERT. Specifically, as in Figure \ref{fig:overview}, after sampling an anchor text segment, we place either (1) the contiguous segment from the same document, (2) a random document, or (3) a document linked from anchor segment, as the next segment in the input. We then train the LM with two joint objectives: We use masked language modeling (MLM) to encourage learning multi-hop knowledge of concepts brought into the same context by document links (e.g.~``{Tidal Basin}'' and ``{Japanese cherry}'' in Figure \ref{fig:motivation}). Simultaneously, we propose a Document Relation Prediction (DRP) objective, which classifies the relation of the second segment to the first segment (\textit{contiguous}, \textit{random}, or \textit{linked}).
DRP encourages learning the relevance and bridging concepts (e.g.~``{National Cherry Blossom Festival}'') between documents, beyond the ability learned in the vanilla next sentence prediction objective in BERT.

Viewing the pretraining corpus as a graph of documents, LinkBERT is also motivated as self-supervised learning on the graph, where DRP and MLM correspond to link prediction and node feature prediction in graph machine learning \cite{yang2015embedding,hu2019pretraining}. Our modeling approach thus provides a natural fusion of language-based and graph-based self-supervised learning.

We train LinkBERT in two domains: the general domain, using Wikipedia articles with hyperlinks (\S \ref{sec:method}), and the biomedical domain, using PubMed articles with citation links (\S \ref{sec:biomed}). We then evaluate the pretrained models on a wide range of downstream tasks such as question answering, in both domains. 
LinkBERT consistently improves on baseline LMs across domains and tasks. For the general domain, LinkBERT outperforms BERT on MRQA benchmark (+4\% absolute in F1-score) as well as GLUE benchmark.
For the biomedical domain, LinkBERT exceeds PubmedBERT \cite{gu2020domain} and sets new states of the art on BLURB biomedical NLP benchmark (+3\% absolute in BLURB score) and MedQA-USMLE reasoning task (+7\% absolute in accuracy).
Overall, LinkBERT attains notably large gains for multi-hop reasoning, multi-document understanding, and few-shot question answering, suggesting that LinkBERT internalizes significantly more knowledge than existing LMs by pretraining with document link information. 


%% file: 060_related.tex
\section{Related work}\label{sec:related}

\paragraph{Retrieval-augmented LMs.}
Several works \cite{lewis2020retrieval, karpukhin2020dense, oguz2020unified, xie2022unifiedskg} introduce a retrieval module for LMs, where given an anchor text (e.g.~question), retrieved text is added to the same LM context to improve model inference (e.g.~answer prediction). These works show the promise of placing related documents in the same LM context at inference time, but they do not study the effect of doing so in pretraining. \citet{guu2020realm} pretrain an LM with a retriever that learns to retrieve text for answering masked tokens in the anchor text.
In contrast, our focus is not on retrieval, but on pretraining a general-purpose LM that \textit{internalizes} knowledge that spans across documents, which is orthogonal to the above works (e.g., our pretrained LM could be used to initialize the LM component of these works). Additionally, we focus on incorporating document links such as hyperlinks, which can offer salient knowledge that common lexical retrieval methods may not provide \cite{asai2020learning}.

\paragraph{Pretrain LMs with related documents.}
Several concurrent works use multiple related documents to pretrain LMs. \citet{caciularu2021cross} place documents (news articles) about the same topic into the same LM context, and \citet{levine2021inductive} place sentences of high lexical similarity into the same context.
Our work provides a general method to incorporate document links into LM pretraining, where lexical or topical similarity can be one instance of document links, besides hyperlinks. We focus on hyperlinks in this work, because we find they can bring in salient knowledge that may not be obvious via lexical similarity, and yield a more performant LM (\S \ref{sec:experiment-ablation}).
Additionally, we propose the DRP objective, which improves modeling multiple documents and relations between them in LMs (\S \ref{sec:experiment-ablation}).

\paragraph{Hyperlinks and citation links for NLP.}
Hyperlinks are often used to learn better retrieval models. \citet{chang2020pre, asai2020learning, seonwoo2021weakly} use Wikipedia hyperlinks to train retrievers for open-domain question answering. \citet{ma2021pre} study various hyperlink-aware pretraining tasks for retrieval. While these works use hyperlinks to learn retrievers, we focus on using hyperlinks to create better context for learning general-purpose LMs. Separately, \citet{calixto2021wikipedia} use Wikipedia hyperlinks to learn multilingual LMs.
Citation links are often used to improve summarization and recommendation of academic papers \cite{qazvinian2008scientific, yasunaga2019scisummnet, bhagavatula2018content, khadka2020exploiting, cohan2020specter}. Here we leverage citation networks to improve pretraining general-purpose LMs.

\paragraph{Graph-augmented LMs.}
Several works augment LMs with graphs, typically, knowledge graphs (KGs) where the nodes capture entities and edges their relations. \citet{zhang2019ernie, he2020integrating, wang2021kepler} combine LM training with KG embeddings. \citet{sun2020colake, yasunaga2021qa, zhang2022greaselm} combine LMs and graph neural networks (GNNs) to jointly train on text and KGs. Different from KGs, we use document graphs to learn knowledge that spans across documents.

%% file: 020_preliminary.tex
\section{Preliminaries}\label{sec:prelim}

\noindent
A language model (LM) can be pretrained from a corpus of documents, $\mathcal{X} = \{X^{(i)}\}$. An LM is a composition of two functions, $\fhead(\fenc(X))$, where the encoder $\fenc$ takes in a sequence of tokens $X = (x_1, x_2, ..., x_n)$ and produces a contextualized vector representation for each token, $(\mathbf{h}_1, \mathbf{h}_2, ..., \mathbf{h}_n)$. The head $\fhead$ uses these representations to perform self-supervised tasks in the pretraining step and to perform downstream tasks in the fine-tuning step. We build on BERT \cite{devlin2018bert}, which pretrains an LM with the following two self-supervised tasks.

\heading{Masked language modeling (MLM).}
Given a sequence of tokens $X$, a subset of tokens $Y \subseteq X$ is masked, and the task is to predict the original tokens from the modified input. $Y$ accounts for 15\% of the tokens in $X$; of those, 80\% are replaced with \mask, 10\% with a random token, and 10\% are kept unchanged.

\heading{Next sentence prediction (NSP).}
The NSP task takes two text segments\footnote{A segment is typically a sentence or a paragraph.} $(X_A , X_B)$ as input, and predicts whether $X_B$ is the direct continuation of $X_A$.
Specifically, BERT first samples $X_A$ from the corpus, and then either (1) takes the next segment $X_B$ from the same document, or (2) samples $X_B$ from a random document in the corpus. The two segments are joined via special tokens to form an input instance, \cls $X_A$ \sep $X_B$ \sep, where the prediction target of \cls is whether $X_B$ indeed follows $X_A$ (\textit{contiguous} or \textit{random}).

\paragraph{}
In this work, we will further incorporate document link information into LM pretraining. Our approach (\S \ref{sec:method}) will build on MLM and NSP.

%% file: 030_method.tex
\section{LinkBERT}\label{sec:method}

We present LinkBERT, a self-supervised pretraining approach that aims to internalize more knowledge into LMs using document link information.
Specifically, as shown in Figure \ref{fig:overview}, instead of viewing the pretraining corpus as a set of documents $\mathcal{X} = \{X^{(i)}\}$, we view it as a \emph{graph} of documents, $\mathcal{G} = (\mathcal{X}, \mathcal{E})$, where $\mathcal{E} = \{(X^{(i)}, X^{(j)})\}$ denotes links between documents (\S \ref{sec:method-graph}). The links can be existing hyperlinks, or could be built by other methods that capture document relevance.
We then consider pretraining tasks for learning from document links (\S \ref{sec:method-pretrain}): We create LM inputs by placing linked documents in the same context window, besides the existing options of a single document or random documents. We use the MLM task to learn concepts brought together in the context by document links, and we also introduce the Document Relation Prediction (DRP) task to learn relations between documents.
Finally, we discuss strategies for obtaining informative pairs of linked documents to feed into LM pretraining (\S \ref{sec:method-sampledoc}).

\subsection{Document graph}
\label{sec:method-graph}

Given a pretraining corpus, we link related documents so that the links can bring together knowledge that is not available in single documents. We focus on hyperlinks, e.g., hyperlinks of Wikipedia articles (\S \ref{sec:experiment}) and citation links of academic articles (\S \ref{sec:biomed}).
Hyperlinks have a number of advantages. They provide background knowledge about concepts that the document writers deemed useful---the links are likely to have high precision of relevance, and can also bring in relevant documents that may not be obvious via lexical similarity alone (e.g., in Figure \ref{fig:motivation}, while the hyperlinked article mentions ``{Japanese}'' and ``{Yoshino}'' cherry trees, these words do not appear in the anchor article). Hyperlinks are also ubiquitous on the web and easily gathered at scale \cite{aghajanyan2021htlm}.
To construct the document graph, we simply make a directed edge $(X^{(i)}, X^{(j)})$ if there is a hyperlink from document $X^{(i)}$ to document $X^{(j)}$.

For comparison, we also experiment with a document graph built by lexical similarity between documents. For each document $X^{(i)}$, we use the common TF-IDF cosine similarity metric \cite{chen2017reading, yasunaga2017graph} to obtain top-$k$ documents $X^{(j)}$'s and make edges $(X^{(i)} , X^{(j)})$. We use $k=5$.

\subsection{Pretraining tasks}
\label{sec:method-pretrain}

\paragraph{Creating input instances.}
Several works \cite{gao2021making, levine2021inductive} find that LMs can learn stronger dependencies between words that were shown together in the same context during training, than words that were not. To effectively learn knowledge that spans across documents, we create LM inputs by placing linked documents in the same context window, besides the existing option of a single document or random documents.
Specifically, we first sample an anchor text segment from the corpus (Segment A; $X_A \subseteq X^{(i)}$). For the next segment (Segment B; $X_B$), we either (1) use the contiguous segment from the same document ($X_B \subseteq X^{(i)}$), (2) sample a segment from a random document ($X_B \subseteq X^{(j)}$ where $j \neq i$), or (3) sample a segment from one of the documents linked from Segment A ($X_B \subseteq X^{(j)}$ where $(X^{(i)}, X^{(j)}) \in \mathcal{E}$). We then join the two segments via special tokens to form an input instance: \cls $X_A$ \sep $X_B$ \sep.

\paragraph{Training objectives.}
To train the LM, we use two objectives. The first is the MLM objective to encourage the LM to learn multi-hop knowledge of concepts brought into the same context by document links. The second objective, which we propose, is Document Relation Prediction (DPR), which classifies the relation $r$ of segment $X_B$ to segment $X_A$ ($r \in \{\textit{contiguous}, \textit{random}, \textit{linked}\}$). By distinguishing \textit{linked} from \textit{contiguous} and \textit{random}, DRP encourages the LM to learn the relevance and existence of bridging concepts between documents, besides the capability learned in the vanilla NSP objective. To predict $r$, we use the representation of \cls token, as in NSP. Taken together, we optimize:
\begin{align}
\mathcal{L} &=\mathcal{L}_{\text{MLM}} +\mathcal{L}_{\text{DRP}} \\
&= \scalebox{0.97}{\text{$ -\displaystyle\sum_{i} \log~ p(x_{i} \mid \mathbf{h}_{i}) - \log~ p(r \mid \mathbf{h}_{\texttt{\cls}}) $}}
\end{align}
where $x_i$ is each token of the input instance, \cls $X_A$ \sep $X_B$ \sep, and $\mathbf{h}_i$ is its representation.

\paragraph{Graph machine learning perspective.}
Our two pretraining tasks, MLM and DRP, are also motivated as graph self-supervised learning on the document graph. In graph self-supervised learning, two types of tasks, node feature prediction and link prediction, are commonly used to learn the content and structure of a graph. In node feature prediction \cite{hu2019pretraining}, some features of a node are masked, and the task is to predict them using neighbor nodes. This corresponds to our MLM task, where masked tokens in Segment A can be predicted using Segment B (a linked document on the graph), and vice versa. In link prediction \cite{bordes2013translating,wang2021relational}, the task is to predict the existence or type of an edge between two nodes.
This corresponds to our DRP task, where we predict if the given pair of text segments are linked (edge), contiguous (self-loop edge), or random (no edge).
Our approach can be viewed as a natural fusion of language-based (e.g.~BERT) and graph-based self-supervised learning.

\subsection{\scalebox{1}[1]{Strategy to obtain linked documents}}
\label{sec:method-sampledoc}
As described in \S \ref{sec:method-graph}, \S \ref{sec:method-pretrain}, our method \emph{builds} links between documents, and for each anchor segment, \emph{samples} a linked document to put together in the LM input. Here we discuss three key axes to consider to obtain useful linked documents in this process.

\paragraph{Relevance.}
Semantic relevance is a requisite when building links between documents. If links were randomly built without relevance, LinkBERT would be same as BERT, with simply two options of LM inputs (\textit{contiguous} or \textit{random}).
Relevance can be achieved by using hyperlinks or lexical similarity metrics, and both methods yield substantially better performance than using random links (\S \ref{sec:experiment-ablation}).

\paragraph{Salience.}
Besides relevance, another factor to consider (\textit{salience}) is whether the linked document can offer new, useful knowledge that may not be obvious to the current LM. Hyperlinks are potentially more advantageous than lexical similarity links in this regard: LMs are shown to be good at recognizing lexical similarity \cite{zhang2020bertscore}, and hyperlinks can bring in useful background knowledge that may not be obvious via lexical similarity alone \cite{asai2020learning}. Indeed, we empirically find that using hyperlinks yields a more performant LM (\S \ref{sec:experiment-ablation}).

\paragraph{Diversity.}
In the document graph, some documents may have a very high in-degree (e.g., many incoming hyperlinks, like the ``{United States}'' page of Wikipedia), and others a low in-degree. If we {uniformly} sample from the linked documents for each anchor segment, we may include documents of high in-degree too often in the overall training data, losing diversity. To adjust so that all documents appear with a similar frequency in training, we sample a linked document with probability inversely proportional to its in-degree, as done in graph data mining literature \cite{henzinger2000near}. We find that this technique yields a better LM performance (\S \ref{sec:experiment-ablation}).

%% file: 040_experiment.tex
\section{Experiments}\label{sec:experiment}
We experiment with our proposed approach in the general domain first, where we pretrain LinkBERT on Wikipedia articles with hyperlinks (\S \ref{sec:experiment-pretrain}) and evaluate on a suite of downstream tasks (\S \ref{sec:experiment-eval}). We compare with BERT \cite{devlin2018bert} as our baseline.
We experiment in the biomedical domain in \S \ref{sec:biomed}.

\subsection{Pretraining setup}
\label{sec:experiment-pretrain}

\paragraph{Data.} We use the same pretraining corpus used by BERT: Wikipedia and BookCorpus \cite{Zhu_2015_ICCV}. For Wikipedia, we use the WikiExtractor\footnote{\url{https://github.com/attardi/wikiextractor}} to extract hyperlinks between Wiki articles.
We then create training instances by sampling \textit{contiguous}, \textit{random}, or \textit{linked} segments as described in \S \ref{sec:method}, with the three options appearing uniformly (33\%, 33\%, 33\%).
For BookCorpus, we create training instance by sampling \textit{contiguous} or \textit{random} segments (50\%, 50\%) as in BERT.
We then combine the training instances from Wikipedia and BookCorpus to train LinkBERT.
In summary, our pretraining data is the same as BERT, except that we have hyperlinks between Wikipedia articles.

\paragraph{Implementation.}
We pretrain LinkBERT of three sizes, -tiny, -base and -large, following the configurations of BERT\tinymodel (4.4M parameters), BERT\basemodel (110M params), and BERT\largemodel (340M params) \cite{devlin2018bert,turc2019well}.
We use -tiny mainly for ablation studies.

For -tiny, we pretrain from scratch with random weight initialization. We use the AdamW \cite{loshchilov2017decoupled} optimizer with $(\beta_1, \beta_2) = (0.9, 0.98)$, warm up the learning rate for the first 5,000 steps and then linearly decay it. We train for 10,000 steps with a peak learning rate 5e-3, weight decay 0.01, and batch size of 2,048 sequences with 512 tokens. Training took 1 day on two GeForce RTX 2080 Ti GPUs with fp16.

For -base, we initialize LinkBERT with the BERT\basemodel checkpoint released by \citet{devlin2018bert}
and continue pretraining. We use a peak learning rate 3e-4 and train for 40,000 steps.
Other training hyperparameters are the same as -tiny. Training took 4 days on four A100 GPUs with fp16.

For -large, we follow the same procedure as -base, except that we use a peak learning rate of 2e-4. Training took 7 days on eight A100 GPUs with fp16.

\paragraph{Baselines.}
We compare LinkBERT with BERT.
Specifically, for the -tiny scale, we compare with BERT\tinymodel, which we pretrain from scratch with the same hyperparameters as LinkBERT\tinymodel. The only difference is that LinkBERT uses document links to create LM inputs, while BERT does not.

For -base scale, we compare with BERT\basemodel, for which we take the BERT\basemodel release by \citet{devlin2018bert} and continue pretraining it with the vanilla BERT objectives on the same corpus for the same number of steps as LinkBERT\basemodel.

For -large, we follow the same procedure as -base.

\subsection{Evaluation tasks}
\label{sec:experiment-eval}

\input{tbl_mrqa_main}

\input{tbl_mrqa_distract}
\input{tbl_mrqa_0.1}
\input{tbl_ablation}

We fine-tune and evaluate LinkBERT on a suite of downstream tasks.

\paragraph{Extractive question answering (QA).}
Given a document (or set of documents) and a question as input, the task is to identify an answer span from the document.
We evaluate on six popular datasets from the MRQA shared task \cite{fisch2019mrqa}:
\textit{HotpotQA} \cite{yang2018hotpotqa},
\textit{TriviaQA} \cite{joshi2017triviaqa},
\textit{NaturalQ} \cite{kwiatkowski2019natural},
\textit{SearchQA} \cite{dunn2017searchqa},
\textit{NewsQA} \cite{trischler2016newsqa}, and
\textit{SQuAD} \cite{rajpurkar2016squad}.
As the MRQA shared task does not have a public test set, we split the dev set in half to make new dev and test sets.
We follow the fine-tuning method BERT \cite{devlin2018bert} uses for extractive QA. More details are provided in Appendix \ref{sec:appendix-finetune}.

\paragraph{GLUE.}
The General Language Understanding Evaluation (GLUE) benchmark \cite{wang2018glue} is a popular suite of sentence-level classification tasks. Following BERT, we evaluate on
\textit{CoLA} \cite{warstadt2019neural},
\textit{SST-2} \cite{socher2013recursive},
\textit{MRPC} \cite{dolan2005automatically},
\textit{QQP},
\textit{STS-B} \cite{cer2017semeval},
\textit{MNLI} \cite{williams2017broad},
\textit{QNLI} \cite{rajpurkar2016squad}, and
\textit{RTE} \cite{dagan2005pascal, haim2006second, giampiccolo2007third}, and report the average score. More fine-tuning details are provided in Appendix \ref{sec:appendix-finetune}.

\subsection{Results}
\label{sec:experiment-result}

Table \ref{tab:mrqa_main} shows the performance (F1 score) on MRQA datasets. LinkBERT substantially outperforms BERT on all datasets. On average, the gain is +4.1\% absolute for the BERT\tinymodel scale, +2.6\% for the BERT\basemodel scale, and +2.5\% for the BERT\largemodel scale. Table \ref{tab:glue} shows the results on GLUE, where LinkBERT performs moderately better than BERT. These results suggest that LinkBERT is especially effective at learning knowledge useful for QA tasks (e.g.~world knowledge), while keeping performance on sentence-level language understanding.

\subsection{Analysis}
\label{sec:experiment-analysis}

We further study when LinkBERT is especially useful in downstream tasks.

\begin{figure}[!t]
    \centering
    \includegraphics[width=0.46\textwidth]{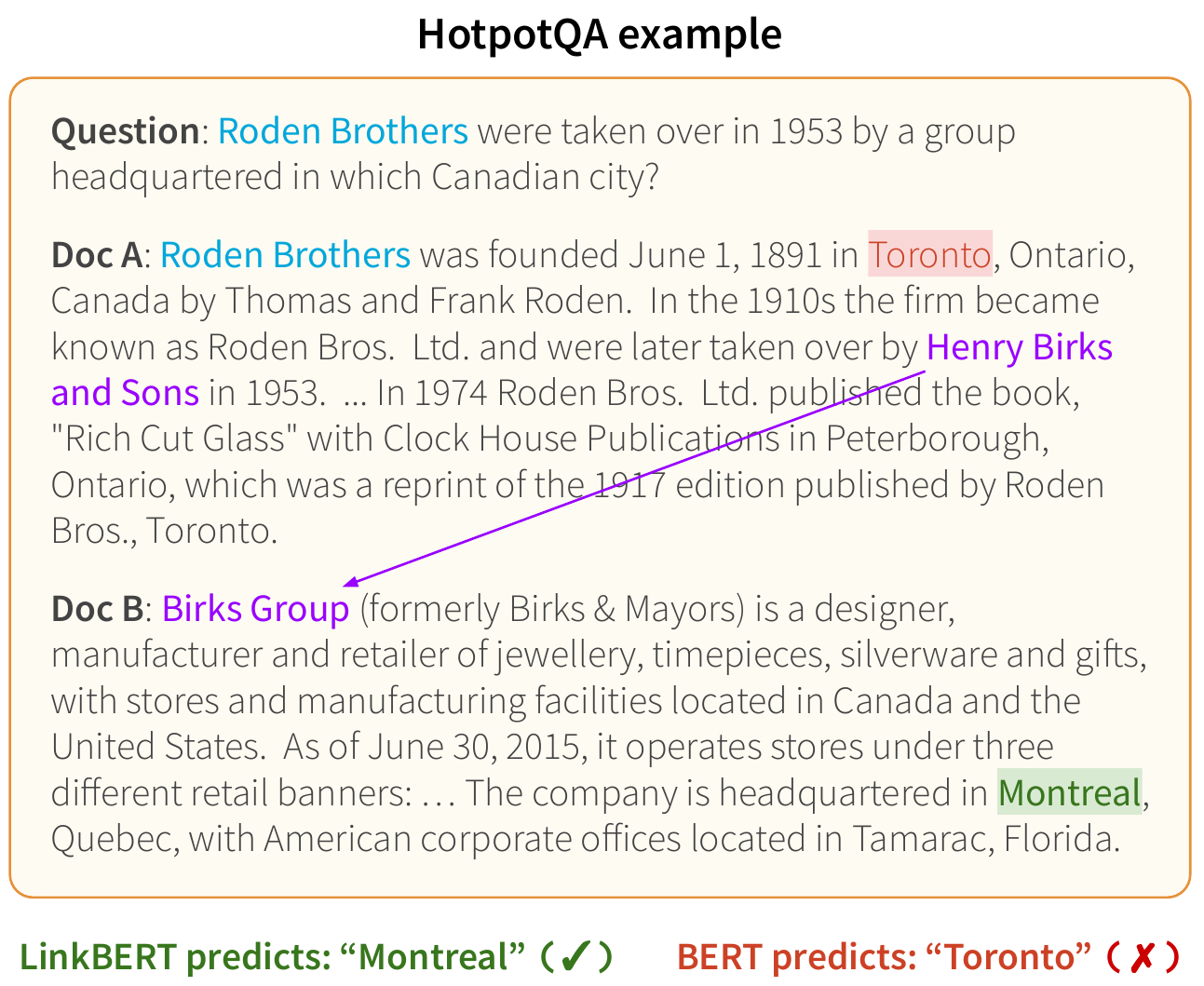}\vspace{-1mm}
    \caption{\small
    Case study of multi-hop reasoning on HotpotQA. Answering the question needs to identify ``{Roden Brothers were taken over by Birks Group}'' from the first document, and then ``{Birks Group is headquartered in Montreal}'' from the second document. While BERT tends to simply predict an entity near the question entity (``{Toronto}'' in the first document), LinkBERT correctly predicts the answer in the second document (``{Montreal}'').
    }
    \vspace{-1mm}
  \label{fig:hotpot}
\end{figure}

\paragraph{Improved multi-hop reasoning.}
In Table \ref{tab:mrqa_main}, we find that LinkBERT obtains notably large gains on QA datasets that require reasoning with multiple documents, such as HotpotQA (+5\% over BERT\tinymodel), TriviaQA (+6\%) and SearchQA (+8\%), as opposed to SQuAD (+1.4\%) which just has a single document per question.
To further gain qualitative insights, we studied in what QA examples LinkBERT succeeds but BERT fails. Figure \ref{fig:hotpot} shows a representative example from HotpotQA. Answering the question needs 2-hop reasoning: identify ``{Roden Brothers were taken over by Birks Group}'' from the first document, and then ``{Birks Group is headquartered in Montreal}'' from the second document. While BERT tends to simply predict an entity near the question entity (``{Toronto}'' in the first document, which is just 1-hop), LinkBERT correctly predicts the answer in the second document (``{Montreal}'').
Our intuition is that because LinkBERT is pretrained with pairs of linked documents rather than purely single documents, it better learns how to flow information (e.g., do attention) across tokens when multiple related documents are given in the context.
In summary, these results suggest that pretraining with linked documents helps for multi-hop reasoning on downstream tasks.

\paragraph{Improved understanding of document relations.}
While the MRQA datasets typically use ground-truth documents as context for answering questions, in open-domain QA, QA systems need to use documents obtained by a retriever, which may include noisy documents besides gold ones \cite{chen2017reading, dunn2017searchqa}. In such cases, QA systems need to understand the document relations to perform well \cite{yang2018hotpotqa}. To simulate this setting, we modify the SQuAD dataset by prepending or appending 1--2 distracting documents to the original document given to each question. Table \ref{tab:mrqa_dist} shows the result.
While BERT incurs a large performance drop (-2.8\%), LinkBERT is robust to distracting documents (-0.5\%). This result suggests that pretraining with document links improves the ability to understand document relations and relevance.
In particular, our intuition is that the DRP objective helps the LM to better recognize document relations like (anchor document, linked document) in pretraining, which helps to recognize relations like (question, right document) in downstream QA tasks. We indeed find that ablating the DRP objective from LinkBERT hurts performance (\S \ref{sec:experiment-ablation}).
The strength of understanding document relations also suggests the promise of applying LinkBERT to various retrieval-augmented methods and tasks (e.g.~\citealt{lewis2020retrieval}), either as the main LM or the dense retriever component.

\paragraph{Improved few-shot QA performance.}
We also find that LinkBERT is notably good at few-shot learning. Concretely, for each MRQA dataset, we fine-tune with only 10\% of the available training data, and report the performance in Table \ref{tab:mrqa_0.1}. In this few-shot regime, LinkBERT attains more significant gains over BERT, compared to the full-resource regime in Table \ref{tab:mrqa_main} (on NaturalQ, 5.4\% vs 1.8\% absolute in F1, or 15\% vs 7\% in relative error reduction).
This result suggests that LinkBERT internalizes more knowledge than BERT during pretraining, which supports our core idea that document links can bring in new, useful knowledge for LMs.

\subsection{Ablation studies}
\label{sec:experiment-ablation}
We conduct ablation studies on the key design choices of LinkBERT.

\paragraph{What linked documents to feed into LMs?}
We study the strategies discussed in \S \ref{sec:method-sampledoc} for obtaining linked documents: relevance, salience, and diversity. Table \ref{tab:ablation-context} shows the ablation result on MRQA datasets.
First, if we ignore relevance and use random document links instead of hyperlinks, we get the same performance as BERT (-4.1\% on average; ``random'' in Table \ref{tab:ablation-context}).
Second, using lexical similarity links instead of hyperlinks leads to 1.8\% performance drop (``TF-IDF''). Our intuition is that hyperlinks can provide more salient knowledge that may not be obvious from lexical similarity alone. Nevertheless, using lexical similarity links is substantially better than BERT (+2.3\%), confirming the efficacy of placing relevant documents together in the input for LM pretraining.
Finally, removing the diversity adjustment in document sampling leads to 1\% performance drop (``No diversity'').
In summary, our insight is that to create informative inputs for LM pretraining, the linked documents must be semantically relevant and ideally be salient and diverse.

\paragraph{Effect of the DRP objective.}
Table \ref{tab:ablation-drp} shows the ablation result on the DRP objective (\S \ref{sec:method-pretrain}). Removing DRP in pretraining hurts downstream QA performance. The drop is large on tasks with multiple documents (HotpotQA, TriviaQA, and SQuAD with distracting documents). This suggests that DRP facilitates LMs to learn document relations.

%% file: tbl_mrqa_main.tex
\begin{table*}
\begin{minipage}{.69\textwidth}
\centering
\small
  \scalebox{0.9}{
  \begin{tabular}{lccccccc}
    \toprule
     & \!\!{HotpotQA}\!\! & \!\!{TriviaQA}\!\! & \!\!{SearchQA}\!\! & \!\!{NaturalQ}\!\! & \!\!{NewsQA}\!\!  & \!\!{SQuAD}\!\! & Avg. \\
    \midrule
    \!BERT\tinymodel & 49.8 & 43.4 & 50.2 & 58.9 & 41.3  & 56.6 & 50.0 \\
    \!LinkBERT\tinymodel
    & \textbf{54.6} & \textbf{50.0} & \textbf{58.6} & \textbf{60.3} & \textbf{42.8} & \textbf{58.0} &  \textbf{54.1} \\
    \midrule
    \!BERT\basemodel
    & 76.0 & 70.3 & 74.2 & 76.5 &  65.7 & 88.7 & 75.2 \\
    \!LinkBERT\basemodel
    & \textbf{78.2} & \textbf{73.9} & \textbf{76.8} & \textbf{78.3} & \textbf{69.3} & \textbf{90.1} & \textbf{77.8}  \\
    \midrule
    \!BERT\largemodel & 78.1 & 73.7 & 78.3 & 79.0 & 70.9 & 91.1 & 78.5 \\
    \!LinkBERT\largemodel
    & \textbf{80.8} & \textbf{78.2} & \textbf{80.5} & \textbf{81.0} & \textbf{72.6} & \textbf{92.7} & \textbf{81.0}  \\
    \bottomrule
  \end{tabular}}
  \vspace{-2mm}
  \caption{\small
  Performance (F1) on MRQA question answering datasets. LinkBERT consistently outperforms BERT on all datasets across the -tiny, -base, and -large scales. The gain is especially large on datasets that require reasoning with multiple documents in the context, such as HotpotQA, TriviaQA, SearchQA.
  }
  \label{tab:mrqa_main}
\end{minipage}\hfill
\begin{minipage}{.27\textwidth}
    \centering
    \small
  \scalebox{0.9}{
  \begin{tabular}{lc}
    \toprule
    & {GLUE score}  \\
    \midrule
     \!BERT\tinymodel & 64.3 \\
    \!LinkBERT\tinymodel
    & \textbf{64.6}\\
    \midrule
    \!BERT\basemodel
    & 79.2 \\
    \!LinkBERT\basemodel
    & \textbf{79.6} \\
    \midrule
    \!BERT\largemodel\!\! & 80.7 \\
    \!LinkBERT\largemodel
    & \textbf{81.1} \\
    \bottomrule
  \end{tabular}}
  \vspace{-2mm}
  \caption{\small
  Performance on the GLUE benchmark. LinkBERT attains comparable or moderately improved performance.
  }
  \label{tab:glue}
\end{minipage}\vspace{-3mm}
\end{table*}

%% file: tbl_mrqa_distract.tex
\begin{table}[!t]
  \centering
  \small
  \scalebox{0.85}{
  \begin{tabular}{lcc}
    \toprule
     & SQuAD & SQuAD distract \\
    \midrule
    \!BERT\basemodel & 88.7 & 85.9  \\
    \!LinkBERT\basemodel
    & \textbf{90.1} & \textbf{89.6}  \\
    \bottomrule
  \end{tabular}
  }\vspace{-2mm}
  \caption{\small
  Performance (F1) on SQuAD when distracting documents are added to the context.
  While BERT incurs a large drop in F1, LinkBERT does not, suggesting its robustness in understanding document relations. 
  }\vspace{-2mm}
  \label{tab:mrqa_dist}
\end{table}

%% file: tbl_mrqa_0.1.tex
\begin{table}[!t]
  \centering
  \small
  \scalebox{0.85}{
  \begin{tabular}{lcccc}
    \toprule
     & \!\!{HotpotQA}\!\! & \!\!{TriviaQA}\!\! & \!\!{NaturalQ}\!\! & \!\!{SQuAD}\!\! \\
    \midrule
    \!BERT\basemodel & 64.8 & 59.2 & 64.8 & 79.6  \\
    \!LinkBERT\basemodel
    & \textbf{70.5} & \textbf{66.0} & \textbf{70.2} & \textbf{82.8}   \\
    \bottomrule
  \end{tabular}
  }\vspace{-2mm}
  \caption{\small
  Few-shot QA performance (F1) when 10\% of fine-tuning data is used. LinkBERT attains large gains, suggesting that it internalizes more knowledge than BERT in pretraining.
  }\vspace{-2mm}
  \label{tab:mrqa_0.1}
\end{table}

%% file: tbl_ablation.tex
\begin{table}[!t]
  \centering
  \small
  \scalebox{0.8}{
  \begin{tabular}{lcccc}
    \toprule
     & \!\!\!{HotpotQA}\!\!\! & \!\!\!{TriviaQA}\!\!\! & \!\!\!{NaturalQ}\!\!\! & \!\!\!{SQuAD}\!\!\! \\
    \midrule
    \!\!LinkBERT\tinymodel
    & \textbf{54.6} & \textbf{50.0} & \textbf{60.3} & \textbf{58.0}  \\
    \!\!No diversity & 53.5 & 48.0 & 60.0 & 57.8\\
    \!\!Change hyperlink to TF-IDF\!\! & 50.0 & 48.2 & 59.6 & 57.6\\
    \!\!Change hyperlink to random\!\! & 49.8 & 43.4 & 58.9 & 56.6\\
    \bottomrule
  \end{tabular}
  }\vspace{-2mm}
  \caption{\small
  Ablation study on what linked documents to feed into LM pretraining (\S \ref{sec:method-sampledoc}).
  }\vspace{-2mm}
  \label{tab:ablation-context}
\end{table}

\begin{table}[!t]
  \centering
  \small
  \scalebox{0.8}{
  \begin{tabular}{lccccc}
    \toprule
     & \!\!{HotpotQA}\!\! & \!\!{TriviaQA}\!\! & \!\!{NaturalQ}\!\! & \!\!{SQuAD}\!\! & \!\!\begin{tabular}{@{}l@{}}\vrule width 0pt depth 0pt height 0pt SQuAD\\[-0.5mm] \hspace{1pt}distract\end{tabular} \!\! \\
    \midrule
    \!\!LinkBERT\basemodel\!\!\!
    & \textbf{78.2} & \textbf{73.9} & \textbf{78.3} & \textbf{90.1} & \textbf{89.6} \\
    \!\!No DRP & 76.5 & 72.5 & 77.0 & 89.3 & 87.0\\
    \bottomrule
  \end{tabular}
  }\vspace{-2mm}
  \caption{\small
  Ablation study on the document relation prediction (DRP) objective in LM pretraining (\S \ref{sec:method-pretrain}).
  }\vspace{-1mm}
  \label{tab:ablation-drp}
\end{table}

%% file: 050_biomed.tex
\section{Biomedical LinkBERT (\textit{BioLinkBERT})}
\label{sec:biomed}

Pretraining LMs on biomedical text is shown to boost performance on biomedical NLP tasks \cite{Beltagy2019SciBERT, lee2020biobert, lewis-etal-2020-pretrained, gu2020domain}. Biomedical LMs are typically trained on PubMed, which contains abstracts and citations of biomedical papers. While prior works only use their raw text for pretraining, academic papers have rich dependencies with each other via citations (references). We hypothesize that incorporating citation links can help LMs learn dependencies between papers and knowledge that spans across them.

With this motivation, we pretrain LinkBERT on PubMed with citation links (\S \ref{sec:biomed-pretrain}), which we term \textit{BioLinkBERT}, and evaluate on biomedical downstream tasks (\S \ref{sec:biomed-eval}). As our baseline, we follow and compare with the state-of-the-art biomedical LM, PubmedBERT \cite{gu2020domain}, which has the same architecture as BERT and is trained on PubMed.

\subsection{Pretraining setup}
\label{sec:biomed-pretrain}

\paragraph{Data.} We use the same pretraining corpus used by PubmedBERT: PubMed abstracts (21GB).\footnote{\url{https://pubmed.ncbi.nlm.nih.gov}. We use papers published before Feb. 2020 as in PubmedBERT.}
We use the Pubmed Parser\footnote{\url{https://github.com/titipata/pubmed_parser}} to extract citation links between articles.
We then create training instances by sampling \textit{contiguous}, \textit{random}, or \textit{linked} segments as described in \S \ref{sec:method}, with the three options appearing uniformly (33\%, 33\%, 33\%).
In summary, our pretraining data is the same as PubmedBERT, except that we have citation links between PubMed articles.

\paragraph{Implementation.}

We pretrain BioLinkBERT of -base size (110M params) from scratch, following the same hyperparamters as the PubmedBERT\basemodel \cite{gu2020domain}. Specifically, we use a peak learning rate 6e-4, batch size 8,192, and train for 62,500 steps. We warm up the learning rate in the first 10\% of steps and then linearly decay it. Training took 7 days on eight A100 GPUs with fp16.

Additionally, while the original PubmedBERT release did not include the -large size, we pretrain BioLinkBERT of the -large size (340M params) from scratch, following the same procedure as -base, except that we use a peak learning rate of 4e-4 and warm up steps of 20\%. Training took 21 days on eight A100 GPUs with fp16.

\paragraph{Baselines.}
We compare BioLinkBERT with PubmedBERT released by \citet{gu2020domain}.

\subsection{Evaluation tasks}
\label{sec:biomed-eval}
For downstream tasks, we evaluate on the BLURB benchmark \cite{gu2020domain}, a diverse set of biomedical NLP datasets, and MedQA-USMLE \cite{jin2021disease}, a challenging biomedical QA dataset.

\paragraph{BLURB}
consists of five named entity recognition tasks, a PICO (population, intervention, comparison, and outcome) extraction task, three relation extraction tasks, a sentence similarity task, a document classification task, and two question answering tasks, as summarized in Table \ref{tab:blurb_main}.
We follow the same fine-tuning method and evaluation metric used by PubmedBERT \cite{gu2020domain}.

\paragraph{MedQA-USMLE}
is a 4-way multi-choice QA task that tests biomedical and clinical knowledge. The questions are from practice tests for the US Medical License Exams (USMLE). The questions typically require multi-hop reasoning, e.g., given patient symptoms, infer the likely cause, and then answer the appropriate diagnosis procedure (Figure \ref{fig:ex-usmle}). We follow the fine-tuning method in \citet{jin2021disease}. More details are provided in Appendix \ref{sec:appendix-finetune}.

\paragraph{MMLU-professional medicine} is a multi-choice QA task that tests biomedical knowledge and reasoning, and is part of the popular MMLU benchmark \cite{hendrycks2021measuring} that is used to evaluate massive language models. We take the BioLinkBERT fine-tuned on the above MedQA-USMLE task, and evaluate on this task without further adaptation.

\input{tbl_blurb}
\input{tbl_medqa}

\begin{figure*}[!t]
    \centering
    \vspace{-5mm}
    \hspace{-1mm}
    \includegraphics[width=0.99\textwidth]{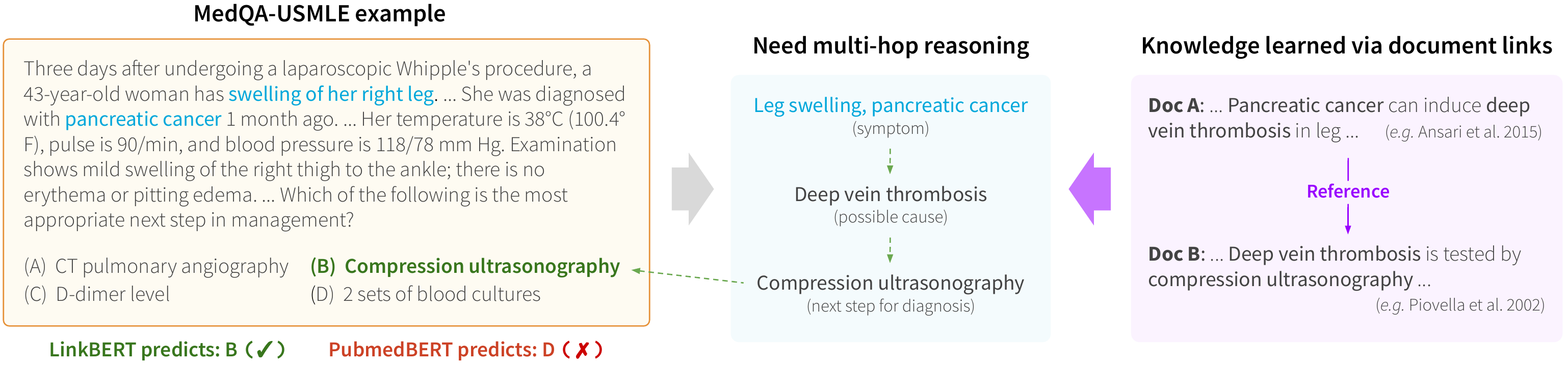}\vspace{-2mm}
    \caption{\small
    Case study of multi-hop reasoning on MedQA-USMLE. Answering the question (left) needs 2-hop reasoning (center): from the patient symptoms described in the question (\textit{leg swelling}, \textit{pancreatic cancer}), infer the cause (\textit{deep vein thrombosis}), and then infer the appropriate diagnosis procedure (\textit{compression ultrasonography}).
    While the existing PubmedBERT tends to simply predict a choice that contains a word appearing in the question (``blood'' for choice D), BioLinkBERT correctly predicts the answer (B).
    Our intuition is that citation links bring relevant documents together in the same context in pretraining (right), which readily provides the multi-hop knowledge needed for the reasoning (center).
    }
    \vspace{-1mm}
  \label{fig:ex-usmle}
\end{figure*}

\subsection{Results}
\paragraph{BLURB.}
Table \ref{tab:blurb_main} shows the results on BLURB. BioLinkBERT\basemodel outperforms PubmedBERT\basemodel on all task categories, attaining a performance boost of {+2}\% absolute on average. Moreover, BioLinkBERT\largemodel provides a further boost of {+1}\%. In total, BioLinkBERT outperforms the previous best by +3\% absolute, establishing a new state of the art on the BLURB leaderboard.
We see a trend that gains are notably large on document-level tasks such as question answering ({+7}\% on BioASQ and PubMedQA). This result is consistent with the general domain (\S \ref{sec:experiment-result}) and confirms that LinkBERT helps to learn document dependencies better.

\paragraph{MedQA-USMLE.}
Table \ref{tab:medqa_main} shows the results.
BioLinkBERT\basemodel obtains a 2\% accuracy boost over PubmedBERT\basemodel, and BioLinkBERT\largemodel provides an additional {+5}\% boost. In total, BioLinkBERT outperforms the previous best by +7\% absolute, setting a new state of the art.
To further gain qualitative insights, we studied in what QA examples BioLinkBERT succeeds but the baseline PubmedBERT fails. Figure \ref{fig:ex-usmle} shows a representative example.
Answering the question (left) needs 2-hop reasoning (center): from the patient symptoms described in the question (\textit{leg swelling}, \textit{pancreatic cancer}), infer the cause (\textit{deep vein thrombosis}), and then infer the appropriate diagnosis procedure (\textit{compression ultrasonography}).
We find that while the existing PubmedBERT tends to simply predict a choice that contains a word appearing in the question (``blood'' for choice D), BioLinkBERT correctly predicts the answer (B).
Our intuition is that citation links bring relevant documents and concepts together in the same context in pretraining (right),\footnote{For instance, as in Figure \ref{fig:ex-usmle} (right), \citet{ansari2015pancreatic} in PubMed mention that \textit{pancreatic cancer can induce deep vein thrombosis in leg}, and it cites a paper in PubMed, \citet{piovella2002normalization}, which mention that \textit{deep vein thrombosis is tested by compression ultrasonography}.
Placing these two documents in the same context yields the complete multi-hop knowledge needed to answer the question (``\textit{pancreatic cancer}'' $\rightarrow$ ``\textit{deep vein thrombosis}'' $\rightarrow$ ``\textit{compression ultrasonography}'').
\vspace{-0mm}} which readily provides the multi-hop knowledge needed for the reasoning (center).
Combined with the analysis on HotpotQA (\S \ref{sec:experiment-analysis}), our results suggest that pretraining with document links consistently helps for multi-hop reasoning across domains (e.g., general documents with hyperlinks and biomedical articles with citation links).

\paragraph{MMLU-professional medicine.} Table \ref{tab:mmlu_main} shows the performance. Despite having just 340M parameters, BioLinkBERT\largemodel achieves 50\% accuracy on this QA task, significantly outperforming the largest general-domain LM or QA models such as GPT-3 175B params (39\% accuracy) and UnifiedQA 11B params (43\% accuracy).
This result shows that with an effective pretraining approach, a small domain-specialized LM can outperform orders of magnitude larger language models on QA tasks.

%% file: tbl_blurb.tex
\begin{table}
\hspace{-1mm}
\small
\scalebox{0.85}{
\begin{tabular}{l @{\vline} cc @{\vline} c}
\toprule
& \begin{tabular}{@{}l@{}}\vrule width 0pt depth 0pt height 8pt \hspace{0.65em}\hspace{0pt}PubMed-\\ \hspace{0.65em}\hspace{0pt}BERT\basemodel\end{tabular}\!\!
& \begin{tabular}{@{}l@{}}\vrule width 0pt depth 0pt height 8pt \hspace{0pt}BioLink-\!\\ \hspace{0pt}BERT\basemodel\hspace{0.3em}~\end{tabular}
& \begin{tabular}{@{}l@{}}\vrule width 0pt depth 0pt height 8pt \hspace{0.7em}\hspace{2pt}BioLink-\!\!\\ \hspace{0.7em}\hspace{0pt}BERT\largemodel\!\!\end{tabular}\!\!\\
\midrule
\!\!\scalebox{0.96}{\textbf{Named entity recognition}} & & \\
~BC5-chem \scalebox{0.7}{\cite{li2016biocreative}} & 93.33 & \textbf{93.75} & \textbf{94.04} \\
~BC5-disease \scalebox{0.7}{\cite{li2016biocreative}} & 85.62 & \textbf{86.10} & \textbf{86.39} \\
~NCBI-disease \scalebox{0.7}{\cite{dougan2014ncbi}} & 87.82 & \textbf{88.18} & \textbf{88.76} \\
~BC2GM \scalebox{0.7}{\cite{smith2008overview}} & {84.52} & \textbf{84.90} & \textbf{85.18} \\
~JNLPBA \scalebox{0.7}{\cite{kim2004introduction}} & \textbf{80.06} & {79.03} & \textbf{80.06} \\
\midrule
\!\!\scalebox{0.96}{\textbf{PICO extraction}} & & \\
~EBM PICO \scalebox{0.7}{\cite{nye2018corpus}} & 73.38 & \textbf{73.97} & \textbf{74.19} \\
\midrule
\!\!\scalebox{0.96}{\textbf{Relation extraction}} & &  \\
~ChemProt \scalebox{0.7}{\cite{krallinger2017overview}} & 77.24 & \textbf{77.57} & \textbf{79.98} \\
~DDI \scalebox{0.7}{\cite{herrero2013ddi}} & 82.36 & \textbf{82.72} & \textbf{83.35}\\
~GAD \scalebox{0.7}{\cite{bravo2015extraction}} & 82.34 & \textbf{84.39} & \textbf{84.90} \\
\midrule
\!\!\scalebox{0.96}{\textbf{Sentence similarity}} & &  \\
~BIOSSES \scalebox{0.7}{\cite{souganciouglu2017biosses}}~~~~ & 92.30 & \textbf{93.25} & \textbf{93.63} \\
\midrule
\!\!\scalebox{0.96}{\textbf{Document classification}} & &  \\
~HoC \scalebox{0.7}{\cite{baker2016automatic}} & 82.32 & \textbf{84.35} & \textbf{84.87}\\
\midrule
\!\!\scalebox{0.96}{\textbf{Question answering}} & &  \\
~PubMedQA \scalebox{0.7}{\cite{jin2019pubmedqa}} & 55.84 & \textbf{70.20} &\textbf{72.18} \\
~BioASQ \scalebox{0.7}{\cite{nentidis2019results}} & 87.56 & \textbf{91.43} & \textbf{94.82} \\
\midrule
\!\!\textbf{BLURB score} &  81.10 & \textbf{83.39} & \textbf{84.30}\\
\bottomrule
\end{tabular}
}\vspace{-2mm}
\caption{\small
{Performance on {BLURB} benchmark.}
BioLinkBERT attains improvement on all tasks, establishing new state of the art on BLURB. Gains are notably large on document-level tasks such as PubMedQA and BioASQ.
}
\label{tab:blurb_main}
\vspace{-2mm}
\end{table}

%% file: tbl_medqa.tex
\begin{table}
\centering
\small
\scalebox{0.9}{
\begin{tabular}{lc}
\toprule
\textbf{Methods}          & \textbf{Acc.} (\%)     \\
\midrule
{BioBERT\largemodel}~ \scalebox{0.9}{\cite{lee2020biobert}}       & 36.7 \\
QAGNN~ \scalebox{0.9}{\cite{yasunaga2021qa}} & 38.0 \\
GreaseLM~ \scalebox{0.9}{\cite{zhang2022greaselm}} & 38.5 \\
\midrule
{PubmedBERT\basemodel}~ \scalebox{0.9}{\cite{gu2020domain}}     & 38.1\\[0.1em]
{BioLinkBERT\basemodel}  (\textbf{Ours})  &  \textbf{40.0}  \\
\midrule
{BioLinkBERT\largemodel}  (\textbf{Ours})  &  \textbf{44.6}  \\
\bottomrule
\end{tabular}
}\vspace{-2mm}
\caption{\small
{Performance on {MedQA-USMLE}}. BioLinkBERT outperforms all previous biomedical LMs.
}
\label{tab:medqa_main}
\vspace{-1mm}
\end{table}

\begin{table}
\centering
\small
\scalebox{0.9}{
\begin{tabular}{lc}
\toprule
\textbf{Methods}          & \textbf{Acc.} (\%)     \\
\midrule
{GPT-3 (175B params)}~ \scalebox{0.9}{\cite{brown2020language}} & 38.7 \\
UnifiedQA (11B params)~ \scalebox{0.9}{\cite{khashabi2020unifiedqa}} & 43.2 \\
\midrule
{BioLinkBERT\largemodel} (\textbf{Ours})  &  \textbf{50.7} \\
\bottomrule
\end{tabular}
}\vspace{-2mm}
\caption{\small
{Performance on {MMLU-professional medicine}}. BioLinkBERT significantly outperforms the largest general-domain LM or QA model, despite having just 340M parameters.
}
\label{tab:mmlu_main}
\vspace{-1mm}
\end{table}

%% file: 070_conclusion.tex
\section{Conclusion}\label{sec:conclusion}
We presented LinkBERT, a new language model (LM) pretraining method that incorporates document link knowledge such as hyperlinks. In both the general domain (pretrained on Wikipedia with hyperlinks) and biomedical domain (pretrained on PubMed with citation links), LinkBERT outperforms previous BERT models across a wide range of downstream tasks. The gains are notably large for multi-hop reasoning, multi-document understanding and few-shot question answering, suggesting that LinkBERT effectively internalizes salient knowledge through document links. Our results suggest that LinkBERT can be a strong pretrained LM to be applied to various knowledge-intensive tasks.

%% file: 080_appendix.tex
\appendix

\section{Ethics, limitations and risks}
\label{sec:appendix-ethics}

We outline potential ethical issues with our work below. First, \methodname is trained on the same text corpora (e.g., Wikipedia, Books, PubMed) as in existing language models. Consequently, \methodname could reflect the same biases and toxic behaviors exhibited by language models, such as biases about race, gender, and other demographic attributes \citep{sheng2020towards}.

Another source of ethical concern is the use of the MedQA-USMLE evaluation \cite{jin2021disease}. While we find this clinical reasoning task to be an interesting testbed for \methodname and for multi-hop reasoning in general, we do not encourage users to use the current models for real world clinical prediction.

\section{Fine-tuning details}
\label{sec:appendix-finetune}

We apply the following fine-tuning hyperparameters to all models, including the baselines.

\paragraph{MRQA.}
For all the extractive question answering datasets, we use \texttt{max\_seq\_length} = 384 and a sliding window of size $128$ if the lengths are longer than \texttt{max\_seq\_length}. 

For the -tiny scale (BERT\tinymodel, LinkBERT\tinymodel), we choose learning rates from \{5e-5, 1e-4, 3e-4\}, batch sizes from \{16, 32, 64\}, and fine-tuning epochs from \{5, 10\}.

For -base (BERT\basemodel, LinkBERT\basemodel), we choose learning rates from \{2e-5, 3e-5\}, batch sizes from \{12, 24\}, and fine-tuning epochs from \{2, 4\}.

For -large (BERT\largemodel, LinkBERT\largemodel), we choose learning rates from \{1e-5, 2e-5\}, batch sizes from \{16, 32\}, and fine-tuning epochs from \{2, 4\}.

\paragraph{GLUE.}
We use \texttt{max\_seq\_length} = 128. 

For the -tiny scale (BERT\tinymodel, LinkBERT\tinymodel), we choose learning rates from \{5e-5, 1e-4, 3e-4\}, batch sizes from \{16, 32, 64\}, and fine-tuning epochs from \{5, 10\}.

For -base and -large (BERT\basemodel, LinkBERT\basemodel, BERT\largemodel, LinkBERT\largemodel), we choose learning rates from \{5e-6, 1e-5, 2e-5, 3e-5, 5e-5\}, batch sizes from \{16, 32, 64\} and fine-tuning epochs from 3--10.

\paragraph{BLURB.}
We use \texttt{max\_seq\_length} = 512 and choose learning rates from \{1e-5, 2e-5, 3e-5, 5e-5, 6e-5\}, batch sizes from \{16, 32, 64\} and fine-tuning epochs from 1--120. 

\paragraph{MedQA-USMLE.}
We use \texttt{max\_seq\_length} = 512 and choose learning rates from \{1e-5, 2e-5, 3e-5\}, batch sizes from \{16, 32, 64\} and fine-tuning epochs from 1--6. 